\documentclass{article}

\usepackage{arxiv}

\usepackage[utf8]{inputenc} 
\usepackage[T1]{fontenc}    
\usepackage{hyperref}       
\usepackage{url}            
\usepackage{booktabs}       
\usepackage{amsfonts}       
\usepackage{nicefrac}       
\usepackage{microtype}      
\usepackage{amsmath} 
\usepackage{graphicx}

\title{STN-Homography: estimate homography parameters directly}

\author{
  Qiang Zhou \\
  \texttt{zhouqiang@zju.edu.cn} \\
  \And
  Xin Li \\
  \texttt{vortexdoctor@zju.edu.cn} \\
}

\begin{document}
\maketitle

\begin{abstract}
In this paper, we introduce the STN-Homography model to directly estimate the homography matrix between image pair. Different most CNN-based homography estimation methods which use an alternative 4-point homography parameterization, we use prove that, after coordinate normalization, the variance of elements of coordinate normalized $3\times3$ homography matrix is very small and suitable to be regressed well with CNN. Based on proposed STN-Homography, we use a hierarchical architecture which stacks several STN-Homography models and successively reduce the estimation error. Effectiveness of the proposed method is shown through experiments on MSCOCO dataset, in which it significantly outperforms the state-of-the-art. The average processing time of our hierarchical STN-Homography with 1 stage is only 4.87 ms on the GPU, and the processing time for hierarchical STN-Homography with 3 stages is 17.85 ms. The code will soon be open sourced.
\end{abstract}


\section{Introduction}

Homologous is a mapping between two images in a plane from different perspectives. They play a vital role in robotics and computer vision applications such as image stitching \cite{Brown2006, Li2018}, monocular SLAM \cite{Mur-Artal2015, Mur-Artal2017}, 3D camera posture reconstruction \cite{Zhang1996, Park2010} and virtual touring \cite{Pan2004}. 

The basic approach to tackle a homography estimation is to use two sets of corresponding points in Direct Linear Transform (DLT) method. However, finding the corresponding set of points from images is not always an easy task. In this regard, there have been significant amount of research. Features such as SIFT \cite{Lowe2004} and ORB \cite{Rublee2011} are used to find the interest points, and employing a matching framework, point correspondences are achieved. Commonly, a RANSAC \cite{Fischler1981} approach is applied on the correspondence set in order to avoid incorrect associations. And, after an iterative optimization process, the best estimate is chosen.

One major problem with such methods such as ORB+RANSAC is their requirements for the hand-crafted features and exhaustive matching step. Convolutional neural network (CNN) automates feature extraction and provides much stronger features than conventional approaches. Their superiority has been shown many times in various tasks \cite{Badrinarayanan2017, He2017, Cao2018, Mikolov2013}. Recently, attempts have been made to solve the problem of matching with CNN. Flownet\cite{Fischer2015} achieves optical flow estimation by using a parallel convolutional network model to independently extract features from each image. A correlation layer is used to locally match extracted features against each other and aggregate them with responses. The expanded feature set is then used in further convolutional layers. Finally, a refinement stage consisting of de-convolutions is used to map optical flow estimates back to the original image coordinates. Flownet 2.0\cite{Ilg2016} was introduced that is using Flownet models as building blocks to create a hierarchical framework to solve the same problem.

Recently, some attempts have been made to tackle the homography estimation with CNN, and acquired higher accuracy than the ORB+RANSAC method. HomographyNet \cite{DeTone2016} defined the homography between two images by relocation of a set of 4 points, also known as 4-point homography parameterization. Their model is based on the VGG’s architecture \cite{Simonyan2014} with 8 convolutional layers, a pooling layer after every 2 convolutions, and 2 fully connected layers with an L2 loss function that results from the difference between predicted and true 4-point coordinate values. The work of \cite{Nowruzi2017} used hierarchy of twin convolutional regression networks to estimate the homography between a pair of images and improved the prediction accuracy of 4-point homography compared with \cite{DeTone2016}. The work of \cite{Nguyen2017} proposed an unsupervised learning algorithm that trains a Deep Convolutional Neural Network to estimate planar homographies and was also based on 4-point homography parameterization. These works all chose the 4-point homography parameterization because the $3\times3$ parameterization $H$ mixes the rotation, translation, scale, and shear components of the homography transformation. The rotation and shear components tend to have a much smaller magnitude than the translation component, and as a result although an error in their values can greatly impact $H$, it will have a small effect on the L2 loss function of the elements of $H$, which is detrimental for training the neural network. The 4-point homography parameterization does not suffer from these problems.

In this paper, we prove that, we can directly estimate the $3\times3$ parameterization $H$ with CNN, rather than the 4-point homography, and can achieve more accurate results. Specifically, we use STN \cite{Jaderberg2015} to estimate the pixel coordinate normalized homography matrix $\overline{H}$, and the small variance in the magnitude of the elements of the normalized $\overline{H}$ makes it easy to train with CNN. 

Our contributions are as follows: (1) We prove that the $3\times3$ homography matrix can be directly learned well with CNN after pixel coordinate normalization. (2) We propose a hierarchical STN-Homography model and achieve more accurate results compared with the state of the art. (3) We propose a sequence STN-Homography model which can be trained end to end and gets superior results than the hierarchical STN-Homography model.

\section{Dataset}

As in \cite{DeTone2016, Nowruzi2017}, we are also using the COCO 2014 dataset (Microsoft Common Objects in Context)\cite{Lin2014}. First, all the images are converted to gray-scale and are down-sampled to a resolution of $320\times240$. To prepare training and test samples, we choose 118000 images from trainval set of COCO 2014 dataset and 10000 images from test set of COCO 2014 dataset. Later, three samples from each image (denoted as image\_a) are generated in order to increase the dataset size. To achieve this, three random rectangles of size $128\times128$, exclude a boundary region of 32 pixels, is chosen from each image. For each rectangle, a random perturbation in the range of 32 pixels is added to each corner point of the rectangle. This provides us with the target 4-point homography values. Target homography is used with the OpenCV library to warp image\_a to get image\_b, where image\_b is the same size as image\_a. Finally, original corner point coordinates are used within the warped image pair (image\_a and image\_b) to extract the warped patches of patch\_a and patch\_b. We then calculate the normalized homography matrix $\overline{H}_{ba}$ with the following equation,

\begin{equation}
  \overline{H}_{ba} = M H_{ba} M^{-1}
  \label{eq:pixel_coor_normalize}
\end{equation}

where $H_{ba}$ is the homography matrix calculated from the previous generated 4-point homography values and $M = \begin{bmatrix}
 \frac{2}{W}& 0 &-1 \\ 
 0&\frac{2}{H}  & -1\\ 
 0& 0 & 1
\end{bmatrix}$ with $W, H$ be the width and height of patch\_b (patch\_a and patch\_b has the same size of $128\times128$ pixels).

We also use the target 4-point homography with the OpenCV library to warp patch\_a to get patch\_a\_t of same size with patch\_a and the patch\_a\_t will be used to calculate L1 pixel-wise photometric loss. The quintuplet data of (patch\_a, patch\_b, $\overline{H}_{ba}$, patch\_a\_t) is our training sample and are fed as inputs to the network. Note that, the prediction of the network is normalized $\overline{H}_{ba}$, and we need to use Eq. \ref{eq:pixel_coor_normalize} to transform and get the non-normalized homography matrix $H_{ba}$.

\begin{figure}
  \centerline{\includegraphics[width=10cm]{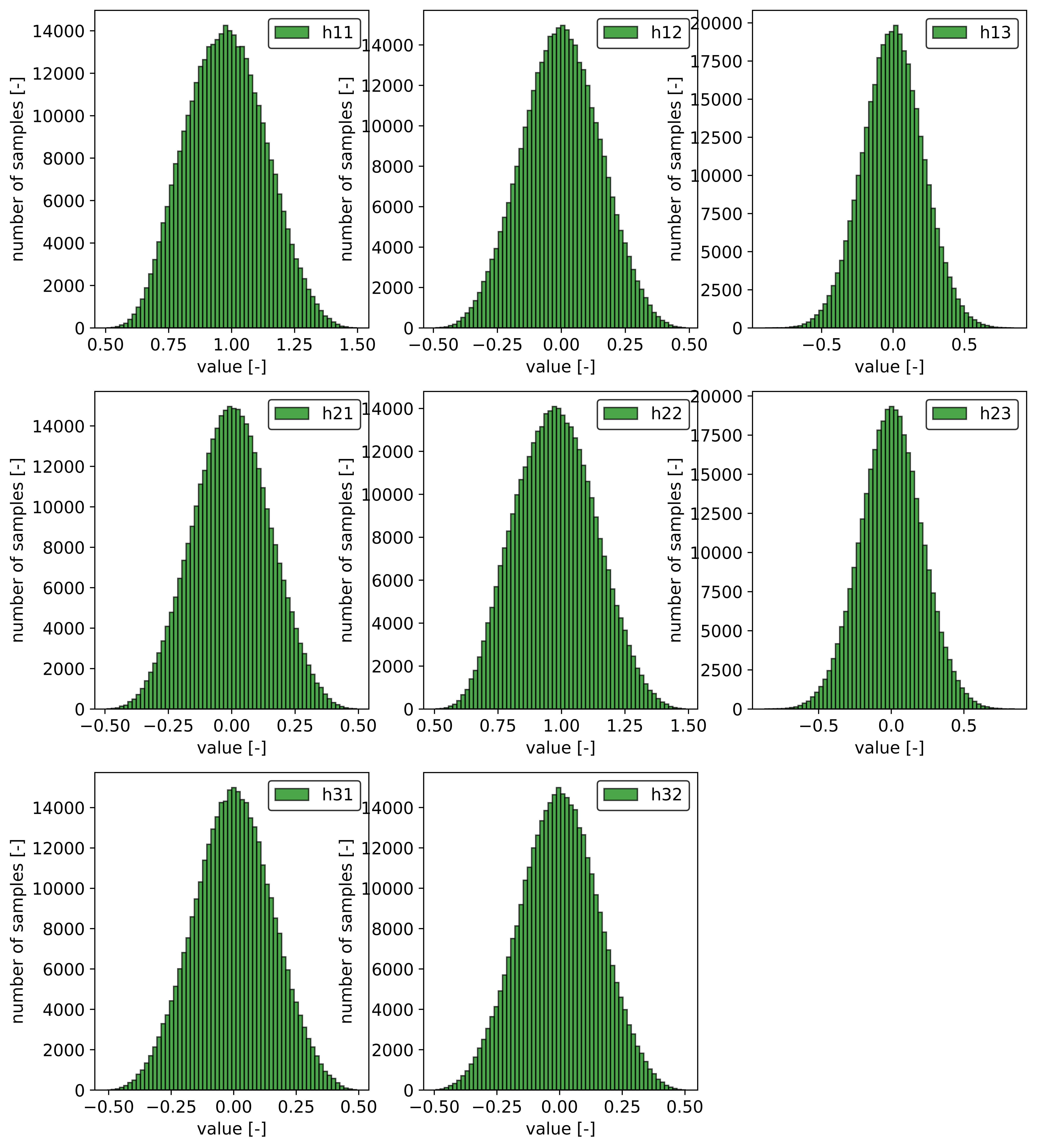}}
  \caption{Value histogram of $\overline{H}_{ba}$ in training dataset.}
  \label{fig:value_histogram}
\end{figure}

Since homography matrix can be multiplied by an arbitrary non-zero scale factor without altering the projective transformation, only the ratio of the matrix elements is significant, leaving $H_{ba}$ ($\overline{H}_{ba}$) eight independent ratios corresponding to eight degree of freedom. And so, we alwasy set the last element of $H_{ba}$ ($\overline{H}_{ba}$) to be 1.0. In the training sample of quintuplet data, we flatten $\overline{H}_{ba}=\begin{bmatrix} h_{11} & h_{12} & h_{13} \\ h_{21} & h_{22} & h_{23} \\ h_{31} & h_{32} & 1.0 \end{bmatrix}$ and take the first eight elements as training input. Fig. \ref{fig:value_histogram} shows the value histogram of $\overline{H}_{ba}$ in training samples after pixel coordinate normalization as depicted in Eq. \ref{eq:pixel_coor_normalize}. From Fig. \ref{fig:value_histogram} we can clearly see that, after normalization, the variance of the eight independent elements of $\overline{H}_{ba}$ is very small, which means $\overline{H}_{ba}$ can be easily regressed with CNN.

\section{STN-Homography Architecture}

\begin{figure}
  \centerline{\includegraphics[width=12cm]{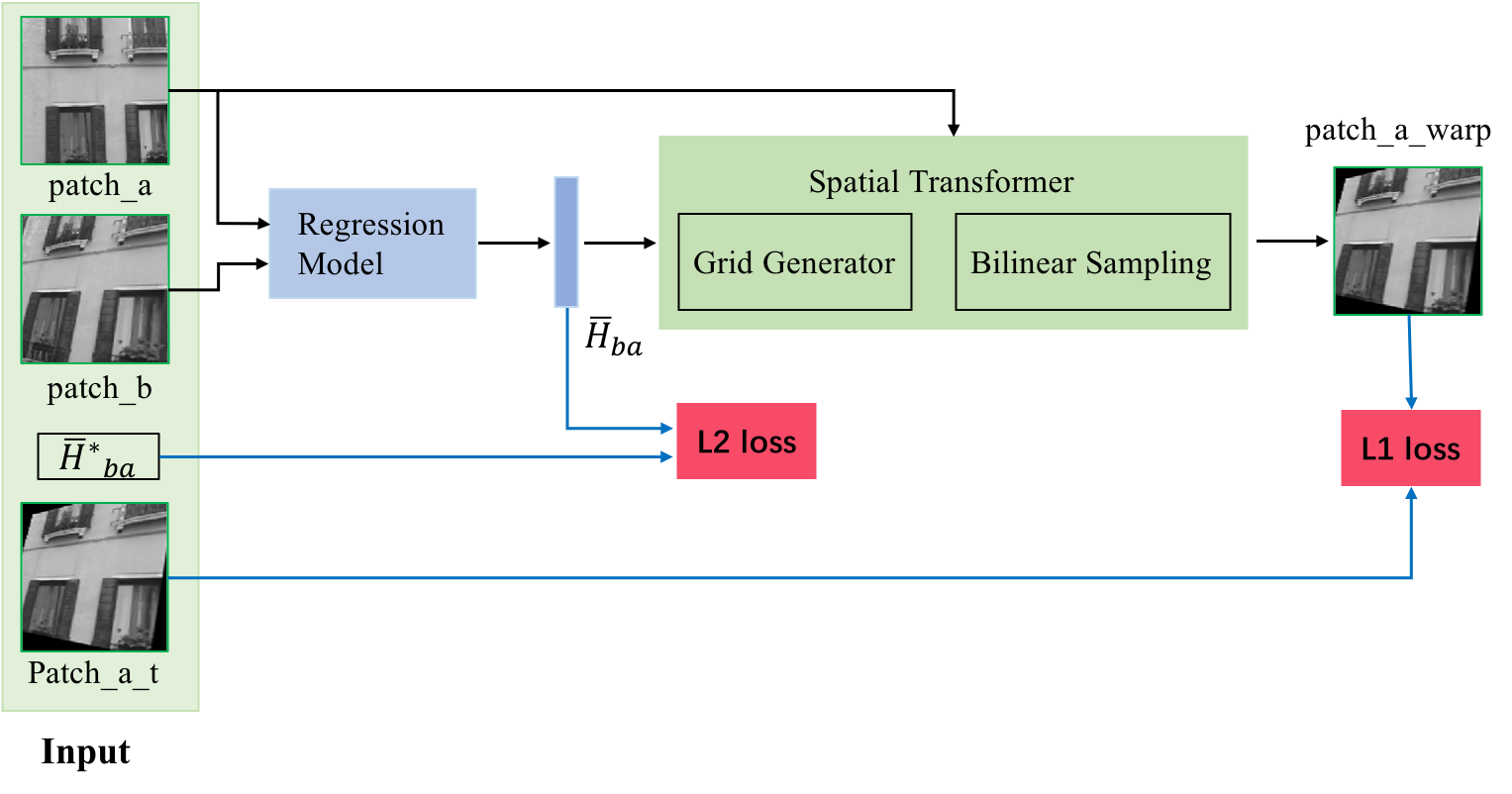}}
  \caption{STN-Homography architecture.}
  \label{fig:structure_stn_homography}
\end{figure}

Fig. \ref{fig:structure_stn_homography} depicts our STN-Homography architecture to predict the pixel coordinate normalized homography matrix $\overline{H}_{ba}$. Our Regression Model outputs 8 regression values corresponding to first 8 elements of flattened $\overline{H}_{ba}$, leaving the last element of $\overline{H}_{ba}$ to be 1. The architecture of our Regression Model is similar to VGG Net \cite{Simonyan2014}. We use 8 convolutional layers with a max pooling layer ($2\times2$, stride 2) after every two convolutions. The 8 convolutional layers have the following number of filters per layer: 64, 64, 64, 64, 128, 128, 128, 128. The output deep features of last convolutional layer are followed by a global average pooling layer, and then two fully connected layers. The first fully connected layer has 1024 units and the second fully connected layer has 8 units. Dropout with a probability of 0.5 is applied after the first fully connected layer. The input to our Regression Model is a two-channel grayscale images sized $128\times128\times2$ pixels. In other words, the two input images of patch\_a and patch\_b, which are related by a homography, are stacked channel-wise and fed into the network.

We use two losses in the training of STN-Homography. The first is L2 loss between the regression output $\overline{H}_{ba}$ and ground true $\overline{H}^*_{ba}$, and the second is the L1 pixel-wise photometric loss between the output of Spatial Transformer and ground truth patch\_a\_t. The same L1 loss is also used in \cite{Nguyen2017}, however \cite{Nguyen2017} is based on 4-point homography while we directly estimate the $3\times3$ homography matrix. With the regression output $\overline{H}_{ba}$, we use differentiable grid generator and bilinear sampling (more detail in \cite{Jaderberg2015}) to warp the patch\_a to get patch\_a\_warp and the compute the L1 loss between patch\_a\_warp and patch\_a\_t. The whole network is differentiable and can be trained with back propagation.

\section{Hierarchical STN-Homography}

\subsection{Architecture of Hierarchical STN-Homography}

\begin{figure}
  \centerline{\includegraphics[width=16cm]{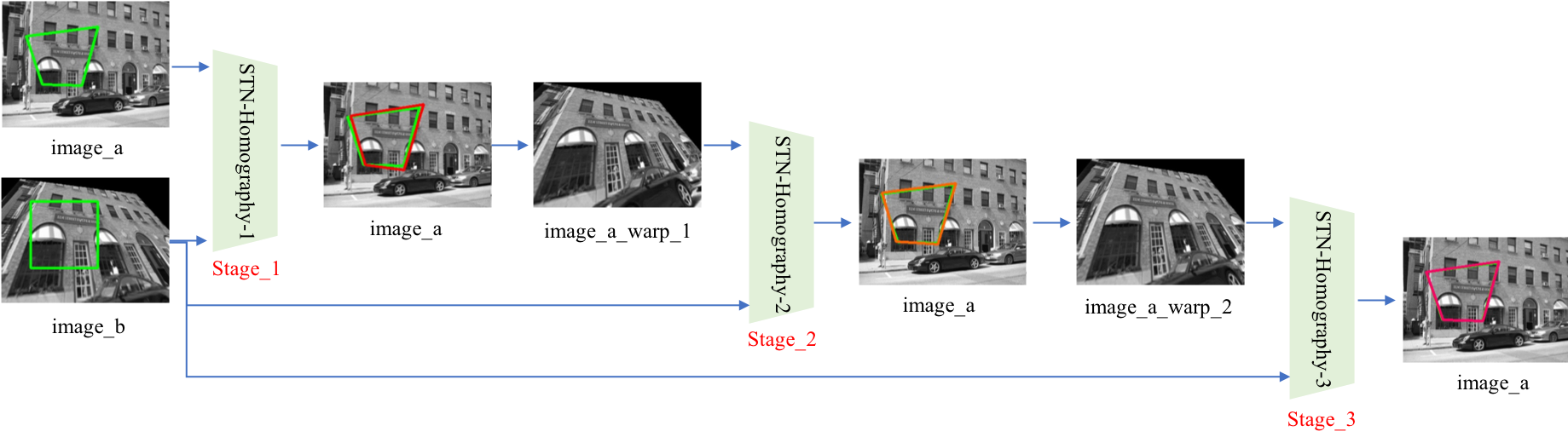}}
  \caption{Hierarchical STN-Homography architecture.}
  \label{fig:structure_hierarchical_stn_homography}
\end{figure}

As same in \cite{Nowruzi2017}, we also use a hierarchical model to successively reduces estimation error, as depicted in Fig. \ref{fig:structure_hierarchical_stn_homography}. In each module, we use a new STN-Homography model to estimate the $\overline{H}_{ba\_i}$ between patch\_a\_i and patch\_b, and the estimated $\overline{H}_{ba\_i}$ will be used with OpenCV library to warp image\_a\_i to prepare image\_a\_i+1 and patch\_a\_i+1 for next module. To calculate the final result that can directly transform one image to another, all homography matrix estimations of successive modules are multiplied togehter.
\begin{equation}
  \overline{H}_{ba} = \prod_{i=n-1, ..., 0} \overline{H}_{ba\_i}
\end{equation}

Warping with predicted homography matrix from each module resulting a visually more similar patch pair (or image pair of image\_a\_warp\_i and image\_a). This can be visualized as a geometric morphing process that takes one image and successively makes it to look alike the other, shown in Fig. \ref{fig:structure_hierarchical_stn_homography}.

\subsection{Training}

As shown in Fig. \ref{fig:structure_hierarchical_stn_homography}, when training the hierarchical modules, the following stage module's training data depends on previous stage module's predictions. If training these three cascade modules at the same time, we need to do some data processing (e.g., warping image\_a to generate image\_a\_warp) on the fly, which will result in very slow training speed. To speed up training, we adopt a step-by-step training strategy and prepare training data for each stage module offline.

For the training of all stage modules, we use the same training parameters for simplicity. Specifically, we use the momentum optimizer with momentum value of 0.9, batch size of 64 and initial learning rate of 0.05. During the first 1000 training steps, we linearly increase the learning rate from 0.0 to the initial learning rate of 0.05. And then, we continue training 90000 steps during which we update the learning rate from 0.05 to 0.0 with cosine decay method \cite{Loshchilov2016}.

\subsection{Accuracy Results}
\begin{figure}
  \centerline{\includegraphics[width=10cm]{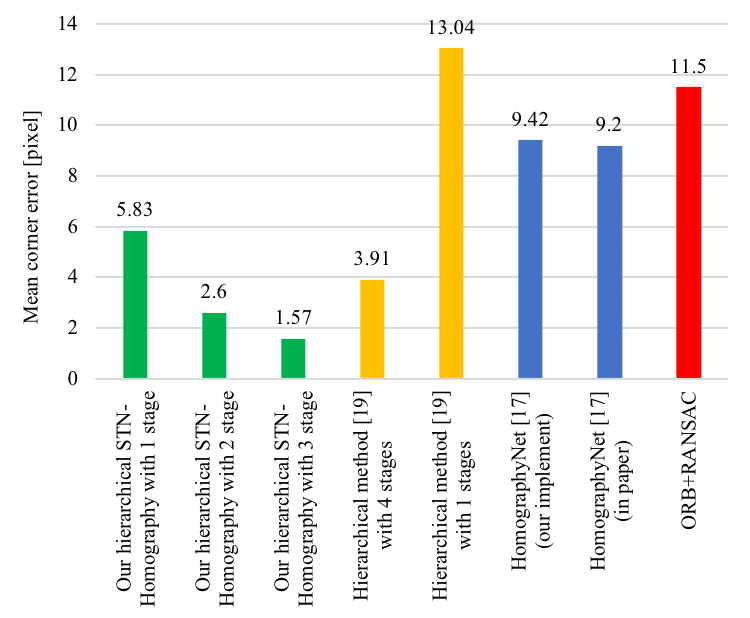}}
  \caption{Mean corner pixel error comparison of various methods for homograph estimation of center aligned image pair}
  \label{fig:mean_corner_err_histgram_center_align}
\end{figure}

First, we experimentally compared the corner error of our hierarchical STN-Homography network with other reported approaches. The corner error is achieved by calculating L2 distance between target and estimated corner locations and averaging them over 4 corners and all test samples. The approaches used for comparison consist of a traditional one and two convolutional one. The selected traditional approach is based on ORB+RANSAC method and the reference deep convolutional approaches are the HomographyNet by \cite{DeTone2016} and hierarchical method by \cite{Nowruzi2017}.

Using our proposed hierarchical STN-Homography, we report in Fig. \ref{fig:mean_corner_err_histgram_center_align} the progressive improvement. As shown in Fig. \ref{fig:mean_corner_err_histgram_center_align}, the mean corner error of our hierarchical STN-Homography with three stages is only 1.57 pixels. When compared with \cite{DeTone2016} of 9.2 pixel error, we decreased the mean corner error by 82.9\%. When compared with the four-stage hierarchical model of \cite{Nowruzi2017}, which has the mean corner error of 3.91 pixels, we also decreased the corner error by 59.8\%.

\subsection{Loss weight comparison}

We use two losses for the training of STN-Homography, i.e., the L2 loss for regressed $\overline{H}_{ba}$ and L1 loss for pixel-wise photometric loss. During the training of previous experiments, we use the same loss weight for these two losses. In this section, we explore the impact of loss weights on our network performance. For simplify, we only use the single STN-Homography model to conduct these experiments. We use the same training parameters for the experiments conducted in Table. \ref{tbl:loss_weight_exp}. 

\begin{table}[!h]
  \centering
  \begin{tabular}{l|c|c|c}
    \hline
    Model name & L2 loss weight & L1 loss weight & mean corner error [pixel] \\
    \hline
    single STN-Homography  & 1.0 & 1.0 & 5.83 \\
    \hline
    single STN-Homography & 1.0 & 10.0 & 21.86 \\
    \hline
    single STN-Homography & 1.0 & 0.1 & 6.21 \\
    \hline
    single STN-Homography & 10.0 & 1.0 & 4.85 \\
    \hline
    single STN-Homography & 0.1 & 1.0 & 6.24 \\
    \hline
  \end{tabular}
  \caption{Impact of loss weights on performation of single STN-Homography model.}
  \label{tbl:loss_weight_exp}
\end{table}

As can be seen from the Table. \ref{tbl:loss_weight_exp}, when the weight of L2 loss remains unchanged, increasing or decreasing the weight of L1 loss will lead to poor accuracy of the STN-Homography. However, if the weight of L1 loss is increased, the accuracy of the model will become rather poor, indicating that L2 loss has a more important impact on the performance of our STN-Homography model. There are two reasons for retaining L1 loss in the our STN-Homography model. One is that the L1 photometric loss can improve the network accuracy (as depicted in Table.\ref{tbl:loss_weight_exp}, when keeping L2 loss weight be 1.0 unchanged and increasing the L1 loss weight from 0.1 to 1.0, the resulting mean corner error is decreased from 6.21 pixels to 5.83 pixels), and the other is that, we can conduct semi-supervised training with the L1 photometric loss to allow some training samples missing ground truth $\overline{H}_{ba}$ (as depicted in \cite{Nguyen2017}, which only use the L1 photometric loss to conduct unsupervised training).

\subsection{Time consumption analysis}

We used Tensorflow \cite{Abadi2016} to implement our proposed network model. During the test time, we achieved an average processing time of 4.87 milli-seconds for a single STN-Homography model on a GPU. When the same STN-Homography model is used in each stage of hierarchical method, the overall computational complexity is given by,

\begin{equation}
  d_e = (l_m + l_w) * n
\end{equation}
where $d_e$ is the end-to-end delay of the whole hierarchical models, $l_m$ is average latency for each STN-Homography model, $l_w$ is the over-head of warping to generate a new image pair, and $n$ is number of stages used in the framework. 

\begin{table}[!h]
  \centering
  \begin{tabular}{l|c}
    \hline
    Model name & Time consumption on the GPU [ms] \\
    \hline
    One-stage hierarchical STN-Homography & 4.87 \\
    \hline
    Two-stage hierarchical STN-Homography & 11.46 \\
    \hline
    Three-stage hierarchical STN-Homography & 17.85 \\
    \hline
  \end{tabular}
  \caption{Time consumption of our hierarchical STN-Homography}
  \label{tbl:time_consuming_hierarchical}
\end{table}

Table. \ref{tbl:time_consuming_hierarchical} shows the time consumption of our hierarchical STN-Homography model on a GPU. It can be seen that, our three-stage hierarchical STN-Homography has an average processing time of 17.85 ms on the GPU. The real-time processing speed satisfies the requirements for most potential applications.

\subsection{Prediction results}

\begin{figure}
  \centerline{\includegraphics[width=16cm]{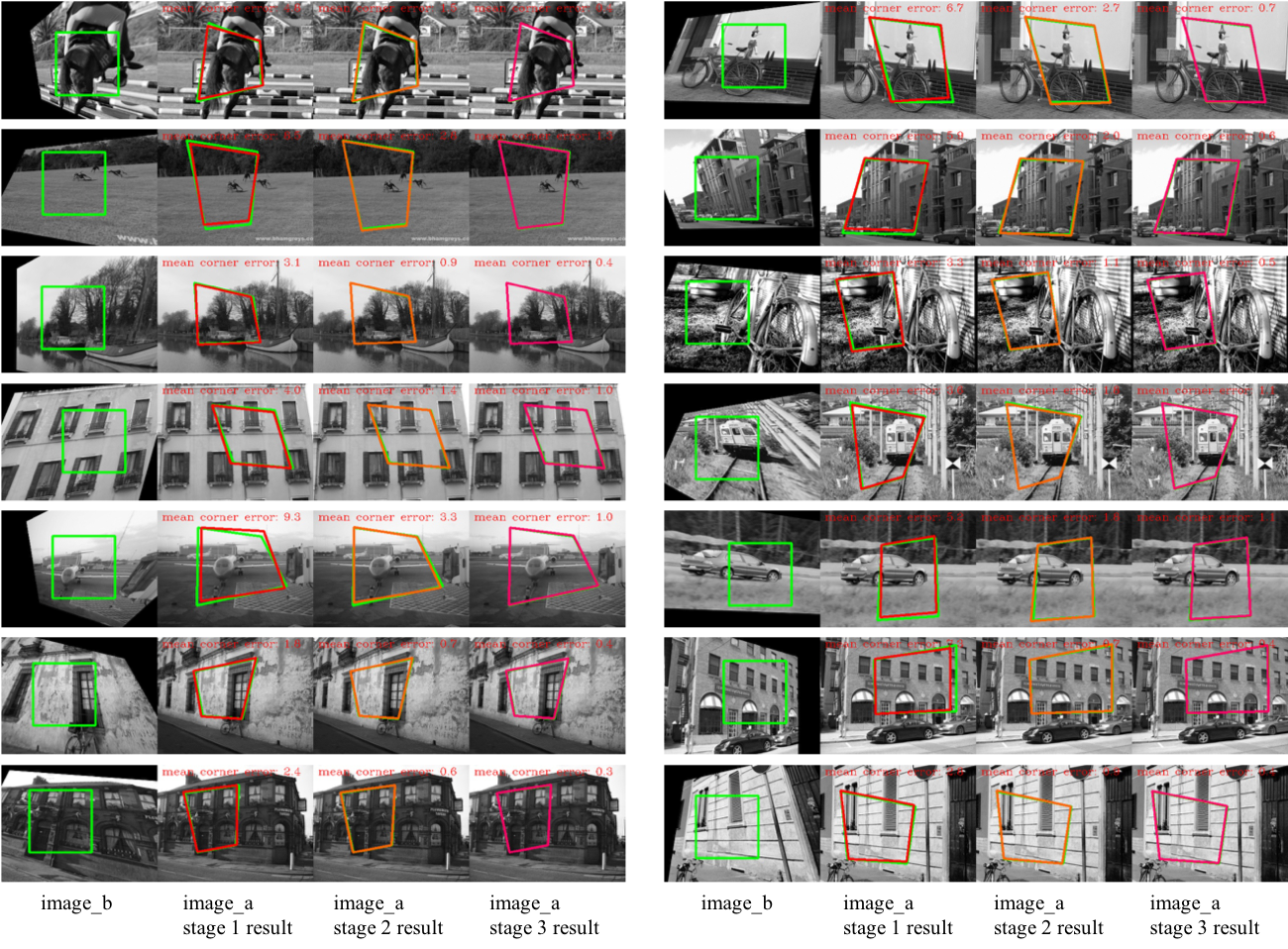}}
  \caption{Prediction results of our three-stage hierarchical STN-Homography.}
  \label{fig:three_stage_hierarchical_pred_res_samples}
\end{figure}

Fig. \ref{fig:three_stage_hierarchical_pred_res_samples} shows the predicted results of our three-stage hierarchical STN-Homography in some test samples. The green boxes in image\_a and image\_b represent the ground truth corresponding points, and the red boxes in image\_a are our predictions. As can be seen from the figure, our model has very small mean corner errors.

\section{Sequence STN-Homography}

Although the previous proposed hierarchical STN-Homography gets very small mean corner error, the training of multi stage hierarchical STN-Homography is not end to end and rely on image\_a when conducting training or test (image\_a was warped using the prediction result of current stage to generate input patch\_a for next stage). In this section, we proposed Sequence STN-Homography which can be trained end to end and did not rely on image\_a when conducting training and test, i.e., Sequential STN-Homography takes input image pair of patch\_a and patch\_b, and directly output the prediction result of homography values.

\subsection{Architecture of Sequence STN-Homography}

\begin{figure}
  \centerline{\includegraphics[width=16cm]{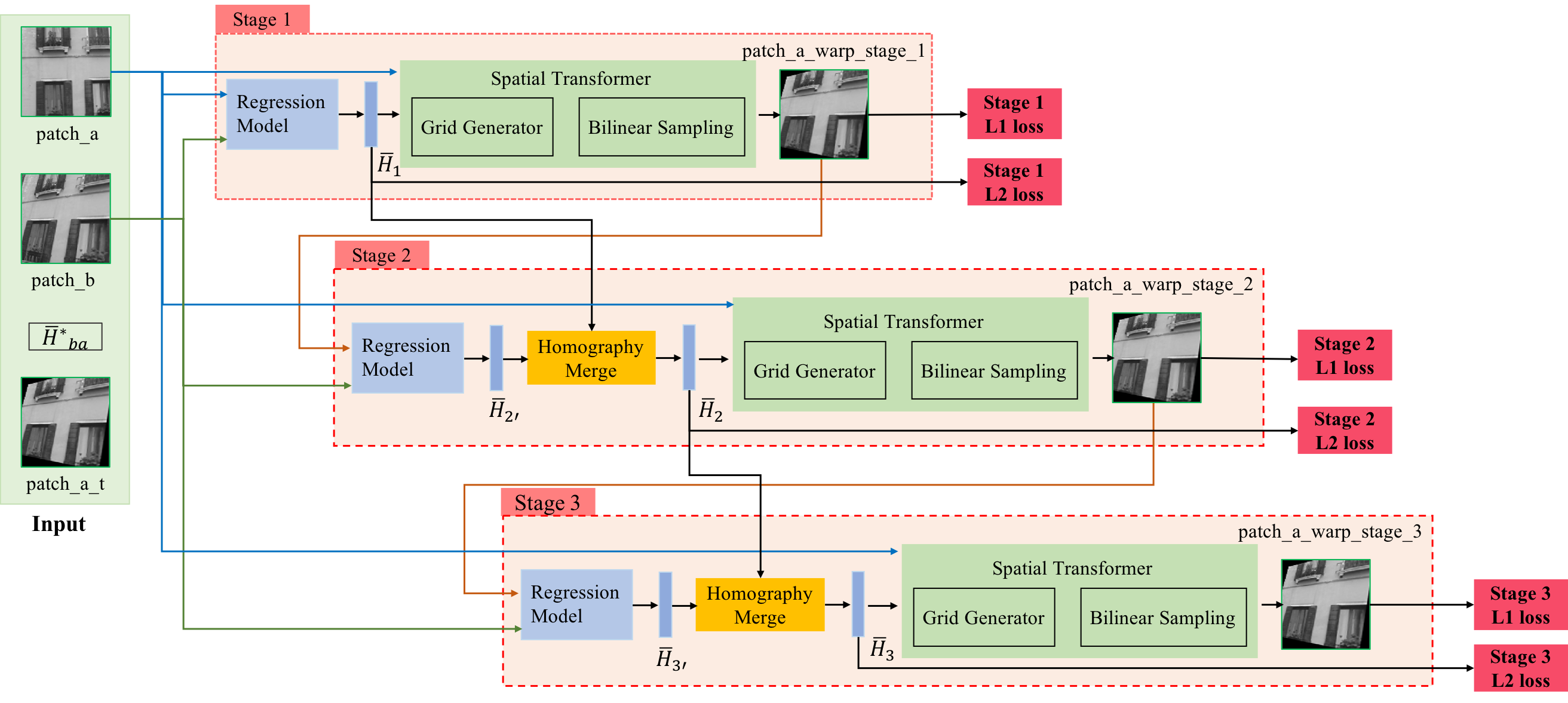}}
  \caption{Architecture of Sequence STN-Homography with 3 stages}
  \label{fig:archi_sequence_stn_homography}
\end{figure}

Sequence STN-Homography is cascaded with several STN-Homography models as depicted in Fig. \ref{fig:archi_sequence_stn_homography}. The training input of sequence STN-Homography is patch\_a, patch\_b, $\overline{H}^*_{ba}$, and patch\_a\_t. Taking there-stage sequence STN-Homography as an example. In stage 1, the STN-Homography model takes input image pair of (patch\_a, patch\_b) and outputs $\overline{H}_1$ and patch\_a\_warp\_stage\_1, where $\overline{H}_1$ is used to compute L2 loss with ground truth $\overline{H}^*_{ba}$ and patch\_a\_warp\_stage\_1 is used to compute the L1 loss with ground truth patch\_a\_t. In stage 2, the STN-Homography model takes input image pair of (patch\_a\_warp\_stage\_1, patch\_b) and outputs $\overline{H}_2$ and patch\_a\_warp\_stage\_2, where $\overline{H}_2$ is used to compute L2 loss with ground truth $\overline{H}^*_{ba}$ and patch\_a\_warp\_stage\_2 is used to compute L1 loss with ground truth patch\_a\_t. In stage 3, the STN-Homography model takes input image pair of (patch\_a\_warp\_stage\_2, patch\_b) and outputs $\overline{H}_3$ and patch\_a\_warp\_stage\_3, where $\overline{H}_3$ is used to compute L2 loss with ground truth $\overline{H}^*_{ba}$ and patch\_a\_warp\_stage\_3 is used to compute L1 loss with ground truth patch\_a\_t.

\begin{equation}
  M^{-1} \overline{H}_1 M I_{\text{patch\_a}} =  I_{\text{patch\_a\_warp\_stage\_1}}
  \label{eq:seq_stage_1}
\end{equation}

\begin{equation}
\begin{matrix}
  M^{-1} \overline{H}_{2'} M I_{\text{patch\_a\_warp\_stage\_1}} = I_{\text{patch\_b}} \\
  M^{-1} \overline{H}_{2} M I_{\text{patch\_a}} = I_{\text{patch\_a\_warp\_stage\_2}}
\end{matrix}
\label{eq:seq_stage_2}
\end{equation}

\begin{equation}
\begin{matrix}
  M^{-1} \overline{H}_{3'} M I_{\text{patch\_a\_warp\_stage\_2}} = I_{\text{patch\_b}} \\
  M^{-1} \overline{H}_{3} M I_{\text{patch\_a}} = I_{\text{patch\_a\_warp\_stage\_3}}
\end{matrix}
\label{eq:seq_stage_3}
\end{equation}

The prediction output $\overline{H}_1$ of stage 1 is the normalized homography matrix between patch\_a and patch\_a\_warp\_stage\_1, as shown in Eq. \ref{eq:seq_stage_1}. In stage 2, $\overline{H}_{2'}$ is the normalized homography between patch\_a\_warp\_stage\_1 and patch\_b, and $\overline{H}_2$ is the normalized homography between patch\_a and patch\_a\_warp\_stage\_2, as shown in Eq. \ref{eq:seq_stage_2}. In stage 3, $\overline{H}_{3'}$ is the normalized homography between patch\_a\_warp\_stage\_2 and patch\_b, and $\overline{H}_3$ is the normalized homography between patch\_a and patch\_a\_warp\_stage\_3, as shown in Eq. \ref{eq:seq_stage_3}. Combing thest equations, we can get $\overline{H}_2 = \overline{H}_{2'} \overline{H}_1$ and $\overline{H}_3 = \overline{H}_{3'} \overline{H}_2$. We develop a Tensor Homography Merge layer to compute $\overline{H}_2$, $\overline{H}_3$ which is differentiable to be trained with back propagation, as show in Fig. \ref{fig:archi_sequence_stn_homography}.

\subsection{Training and Accuracy Results}

\begin{figure}
  \centerline{\includegraphics[width=9cm]{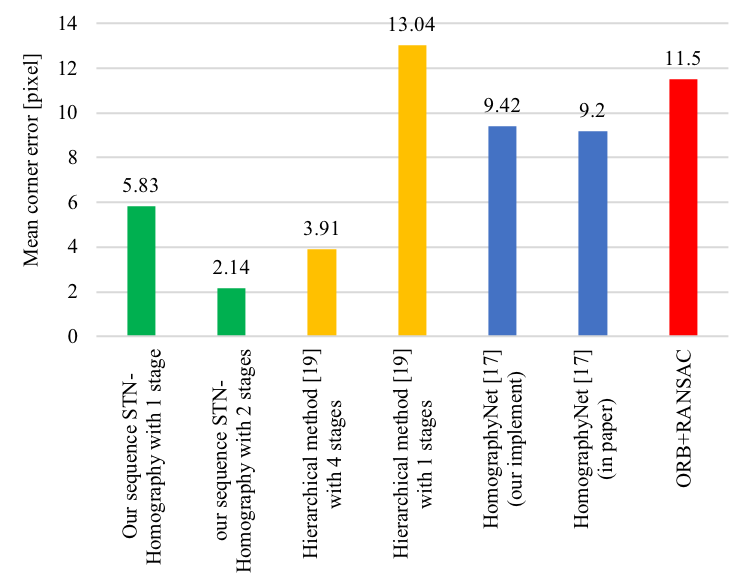}}
  \caption{Mean corner error of our sequence STN-Homography model}
  \label{fig:mean_err_seq_comp}
\end{figure}

For simplify, when training sequence STN-Homography model, we use the same training parameters as used for hierarchical STN-Homography model, i.e., batch size of 64, initial learning rate of 0.05 and total training steps of 90000. Fig. \ref{fig:mean_err_seq_comp} shows the comparison of mean corner error of our sequence STN-Homography model with other reported approaches. At this time, we now only trained two-stage sequence STN-Homography, which achieves the mean corner error of 2.14 pixels, which is lower than the two-stage hierarchical STN-Homography model of 2.6 pixels. We believe that, three-stage sequence STN-Homography model will also be superior than the three-stage hierarchical STN-Homography model.

\section{Conclusion}

In this paper, we have proposed a hierarchical STN-Homography model to target homography estimation. We showed that, after pixel coordinate normalization of homography matrix, we can directly regress the homography matrix values, rather than to estimate the alternative 4-point homography, and the results are significantly better than the state of the art. We use two losses during the training of each STN-Homography model, and find that the L2 loss of $\overline{H}_{ba}$ plays a more import role than the L1 photometric loss in the performance of STN-Homography. While, the use of L1 photometric loss can allowing our model to be trained with a semi-supervised manner, that is, some training samples are allowed to lose ground truth $\overline{H}_{ba}$ values.

\bibliographystyle{unsrt}  


\begin{thebibliography}{10}

\bibitem{Brown2006}
Brown M, Lowe DG.
\newblock Automatic Panoramic Image Stitching using Invariant Features.
\newblock International Journal of Computer Vision. 2006 dec;74(1):59--73.

\bibitem{Li2018}
Li N, Xu Y, Wang C.
\newblock Quasi-Homography Warps in Image Stitching.
\newblock {IEEE} Transactions on Multimedia. 2018 jun;20(6):1365--1375.

\bibitem{Mur-Artal2015}
Mur-Artal R, Montiel JMM, Tardos JD.
\newblock {ORB}-{SLAM}: A Versatile and Accurate Monocular {SLAM} System.
\newblock {IEEE} Transactions on Robotics. 2015 oct;31(5):1147--1163.

\bibitem{Mur-Artal2017}
Mur-Artal R, Tardos JD.
\newblock {ORB}-{SLAM}2: An Open-Source {SLAM} System for Monocular, Stereo,
  and {RGB}-D Cameras.
\newblock {IEEE} Transactions on Robotics. 2017 oct;33(5):1255--1262.

\bibitem{Zhang1996}
Zhang Z, Hanson A.
\newblock 3D Reconstruction Based on Homography Mapping.
\newblock ARPA Image Understanding Workshop. 1996 01;.

\bibitem{Park2010}
Park HS, Shiratori T, Matthews I, Sheikh Y.
\newblock 3D Reconstruction of a Moving Point from a Series of 2D Projections.
\newblock In: Computer Vision {\textendash} {ECCV} 2010. Springer Berlin
  Heidelberg; 2010. p. 158--171.

\bibitem{Pan2004}
Pan Z, Fang X, Shi J, Xu D.
\newblock Easy tour.
\newblock In: Proceedings of the 2004 {ACM} {SIGGRAPH} international conference
  on Virtual Reality continuum and its applications in industry - {ACM} Press; 2004. .

\bibitem{Lowe2004}
Lowe DG.
\newblock Distinctive Image Features from Scale-Invariant Keypoints.
\newblock International Journal of Computer Vision. 2004 nov;60(2):91--110.

\bibitem{Rublee2011}
Rublee E, Rabaud V, Konolige K, Bradski G.
\newblock {ORB}: An efficient alternative to {SIFT} or {SURF}.
\newblock In: 2011 International Conference on Computer Vision. {IEEE}; 2011. .

\bibitem{Fischler1981}
Fischler MA, Bolles RC.
\newblock Random sample consensus: a paradigm for model fitting with
  applications to image analysis and automated cartography.
\newblock Communications of the {ACM}. 1981 jun;24(6):381--395.

\bibitem{Badrinarayanan2017}
Badrinarayanan V, Kendall A, Cipolla R.
\newblock {SegNet}: A Deep Convolutional Encoder-Decoder Architecture for Image
  Segmentation.
\newblock {IEEE} Transactions on Pattern Analysis and Machine Intelligence.
  2017 dec;39(12):2481--2495.

\bibitem{He2017}
He K, Gkioxari G, Dollár P, Girshick R.
\newblock Mask R-CNN;.

\bibitem{Cao2018}
Cao Z, Hidalgo G, Simon T, Wei SE, Sheikh Y.
\newblock OpenPose: Realtime Multi-Person 2D Pose Estimation using Part
  Affinity Fields;.

\bibitem{Mikolov2013}
Mikolov T, Chen K, Corrado G, Dean J.
\newblock Efficient Estimation of Word Representations in Vector Space;.

\bibitem{Fischer2015}
Fischer P, Dosovitskiy A, Ilg E, Häusser P, Hazırbaş C, Golkov V, et~al.
\newblock FlowNet: Learning Optical Flow with Convolutional Networks;.

\bibitem{Ilg2016}
Ilg E, Mayer N, Saikia T, Keuper M, Dosovitskiy A, Brox T.
\newblock FlowNet 2.0: Evolution of Optical Flow Estimation with Deep
  Networks;.

\bibitem{DeTone2016}
DeTone D, Malisiewicz T, Rabinovich A.
\newblock Deep Image Homography Estimation;.

\bibitem{Simonyan2014}
Simonyan K, Zisserman A.
\newblock Very Deep Convolutional Networks for Large-Scale Image Recognition;.

\bibitem{Nowruzi2017}
Nowruzi FE, Laganiere R, Japkowicz N.
\newblock Homography Estimation from Image Pairs with Hierarchical
  Convolutional Networks.
\newblock In: 2017 {IEEE} International Conference on Computer Vision Workshops
  ({ICCVW}). {IEEE}; 2017. .

\bibitem{Nguyen2017}
Nguyen T, Chen SW, Shivakumar SS, Taylor CJ, Kumar V.
\newblock Unsupervised Deep Homography: A Fast and Robust Homography Estimation
  Model;.

\bibitem{Jaderberg2015}
Jaderberg M, Simonyan K, Zisserman A, Kavukcuoglu K.
\newblock Spatial Transformer Networks;.

\bibitem{Lin2014}
Lin TY, Maire M, Belongie S, Bourdev L, Girshick R, Hays J, et~al.
\newblock Microsoft COCO: Common Objects in Context;.

\bibitem{Loshchilov2016}
Loshchilov I, Hutter F.
\newblock SGDR: Stochastic Gradient Descent with Warm Restarts;.

\bibitem{Abadi2016}
Abadi M, Agarwal A, Barham P, Brevdo E, Chen Z, Citro C, et~al.
\newblock TensorFlow: Large-Scale Machine Learning on Heterogeneous Distributed
  Systems;.


\end{thebibliography}

\end{document}